\pdfoutput=1
\documentclass[10pt,twocolumn,letterpaper]{article}

\usepackage{cvpr}              

\usepackage[table]{xcolor}
\usepackage{multirow}

\usepackage{graphicx}
\usepackage{svg}
\usepackage{amsmath}
\usepackage{amssymb}
\usepackage{booktabs}
\usepackage{indentfirst}
\usepackage{tikz}

\usepackage{stfloats}

\definecolor{cvprblue}{rgb}{0.21,0.49,0.74}
\usepackage[pagebackref,breaklinks,colorlinks,citecolor=cvprblue]{hyperref}

\newcommand{\blfootnote}[1]{%
  \begingroup
  \renewcommand\thefootnote{}
  \footnotetext{#1}%
  \endgroup
}

\def\paperID{9710} 
\def\confName{CVPR}
\def\confYear{2025}

\begin{document}

\title{CubeFormer: A Simple yet Effective Baseline for Lightweight \\ Image Super-Resolution}


\author{
Jikai Wang$^{*}$,
Huan Zheng$^{*}$,
Jianbing Shen$^{\dagger}$\\
SKL-IOTSC, CIS, University of Macau
}


\maketitle
\begin{abstract}

Lightweight image super-resolution (SR) methods aim at increasing the resolution and restoring the details of an image using a lightweight neural network.
However, current lightweight SR methods still suffer from inferior performance and unpleasant details.
Our analysis reveals that these methods are hindered by constrained feature diversity, which adversely impacts feature representation and detail recovery.
To respond this issue, we propose a simple yet effective baseline called CubeFormer, designed to enhance feature richness by completing holistic information aggregation.
To be specific, we introduce cube attention, which expands 2D attention to 3D space, facilitating exhaustive information interactions, further encouraging comprehensive information extraction and promoting feature variety.
In addition, we inject block and grid sampling strategies to construct intra-cube transformer blocks (Intra-CTB) and inter-cube transformer blocks (Inter-CTB), which perform local and global modeling, respectively.
Extensive experiments show that our CubeFormer achieves state-of-the-art performance on commonly used SR benchmarks. Our source code and models will be publicly available.

\end{abstract}

\blfootnote{$*$Equal contribution. $\dagger$Corresponding author: \textit{Jianbing Shen}.
}    
\section{Introduction}
\label{sec:intro}

\begin{figure}[t]
    \centering
    \includegraphics[width=1.0 \linewidth]{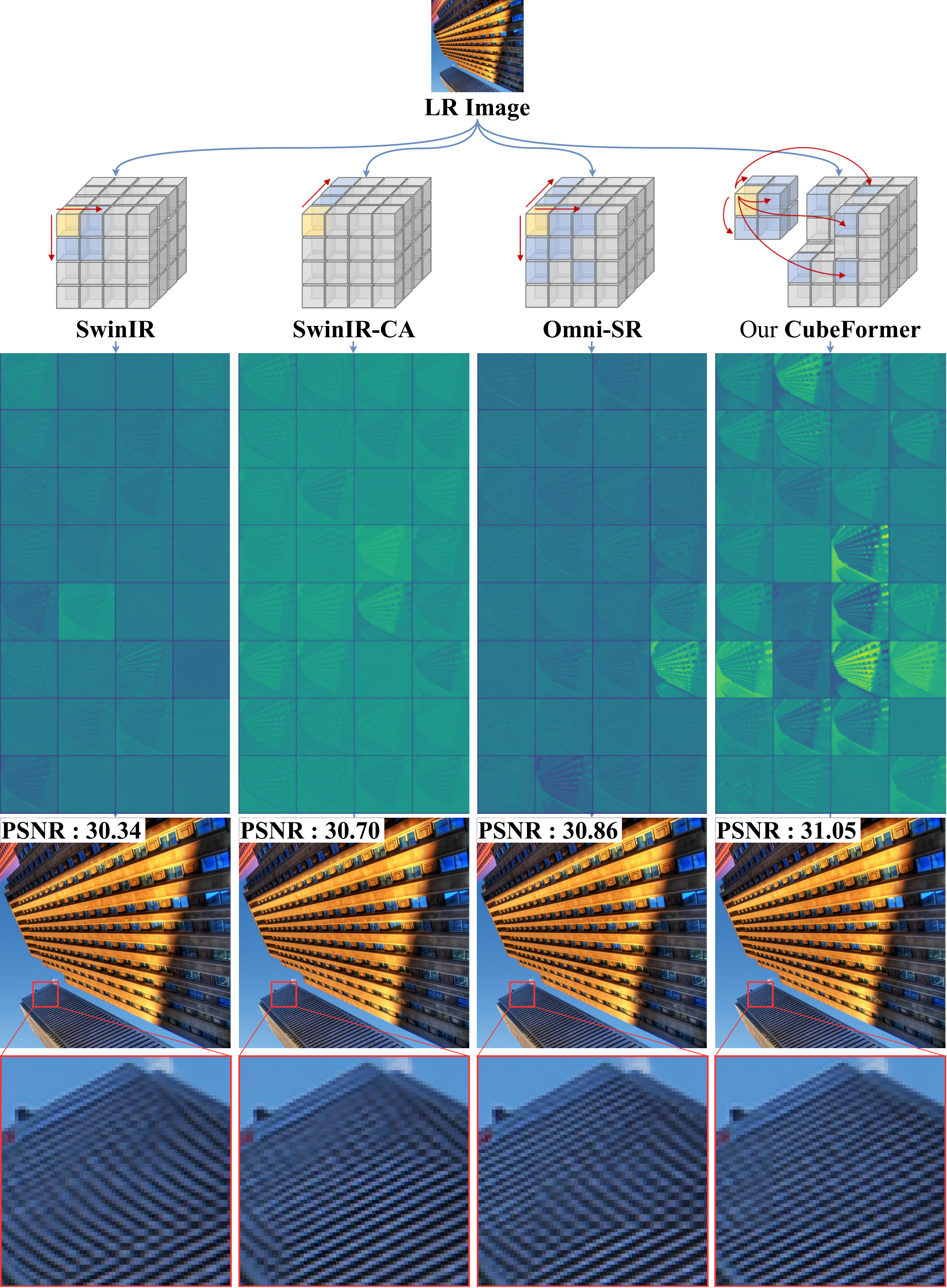}
    \caption{
   \textbf{ An overview of SwinIR~\cite{liang2021swinir}, SwinIR-CA~\cite{zamir2022restormer}, Omni-SR~\cite{wang2023omni}, and the proposed CubeFormer.}
    We visualize the feature maps extracted by the backbone network and restored HR images of these methods. 
    Notably, SwinIR, SwinIR-CA, and Omni-SR capture fewer low-level details in their feature maps, leading to diminished texture clarity in the resulting HR images.
    In contrast, CubeFormer exhibits an enhanced capacity for learning informative representations, effectively enriching fine details in features and delivering HR images of superior quality.
    }
    \vspace{-6mm}
    \label{fig:feat}
\end{figure}

Single-image super-resolution (SR) is a critical task that involves obtaining high-resolution (HR) images from degraded low-resolution (LR) counterparts \cite{dong2015image}. As a long-standing problem, this task has received significant attention, leading to notable advancements in the field \cite{timofte2017ntire, huang2015single, kim2016accurate, zhang2018image}. Some recent works have harnessed the power of the vision transformer (ViT) for image SR \cite{vaswani2017attention, dosovitskiy2020image}. While these ViT-based solutions have demonstrated impressive performance, their deployment often poses a challenge due to their substantial computational demands, primarily stemming from their large model sizes \cite{wang2023global}. Thus, there is a growing imperative to develop lightweight super-resolution models that strike a balance between computational efficiency and performance \cite{wang2023omni, muqeet2020multi}. This paper focuses on designing a lightweight SR framework for better performance.

Current ViT-based lightweight SR methods can be classified into two categories based on the attention mechanisms they employ: homogeneous and heterogeneous structure-based methods. Homogeneous structure-based methods construct a transformer network using either spatial attention \cite{liang2021swinir} or channel attention \cite{zamir2022restormer} to capture correlations. Heterogeneous structure-based methods combine both spatial and channel attentions to aggregate heterogeneous information~\cite{wang2023omni}. 
However, these methods still yield inferior restoration results due to the following limitations: 1) \textit{homogeneous structure-based methods neglect the interaction between spatial and channel information, which is essential for reasoning the details in HR images}; 2) \textit{heterogeneous structure-based methods simply stack two types of attentions, failing to fully exploit the potential low-level features, leading to blurred textures}. 
Figure \ref{fig:feat} shows the comparison and feature visualization of SwinIR~\cite{liang2021swinir}, SwinIR-CA (SwinIR with channel attention) \cite{zamir2022restormer}, Omni-SR~\cite{wang2023omni}, and our CubeFormer. We observed that SwinIR, SwinIR-CA, and Omni-SR suffer from limited feature diversity, indicating limited capacity to capture detailed features. Consequently, the reconstructed HR images lack fine details, as shown by the zoomed-in view. In contrast, CubeFormer can extract more textual features, resulting in the recovered HR image with sharper textures and better detail preservation.

To address the above issue, we propose a simple yet effective lightweight image SR method termed CubeFormer, designed to learn fine-grained textural features, which promotes better detail recovery. 
Specifically, we propose a novel cube attention, dividing all features in 3D space into cubes to enable exhaustive information interactions. 
Consequently, CubeFormer captures omnidirectional relationships and achieves comprehensive information integration, allowing it to retain richer low-level details.
In addition, we introduce two transformer blocks named intra-cube transformer block (Intra-CTB) and inter-cube transformer block (Inter-CTB) to perform simultaneous global and local modeling.
Through the experiments on multiple benchmarks, we demonstrate that the proposed CubeFormer achieves superior performance and detail restoration.

In all, our contributions can be summarized as follows: 
\begin{itemize}
    \item \textbf{CubeFormer}. We propose CubeFormer, a simple yet effective baseline for lightweight image SR. CubeFormer overcomes the limitation of restricted feature diversity in previous solutions and is capable of achieving holistic feature extraction to recover more details.
    \item \textbf{Cube attention}. We develop a novel attention mechanism called cube attention. Cube attention generalizes 2D attention to 3D space, enabling exhaustive interactions and providing enhanced capabilities for detailed feature extraction and reasoning.
    \item \textbf{Intra/inter-cube transformer block}. 
    By embedding block and grid sampling strategies within cube attention, we construct the intra-cube transformer block (Intra-CTB) and inter-cube transformer block (Inter-CTB). These blocks are designed to perform concurrent global and local modeling, which is instrumental in enhancing texture recovery.
    \item \textbf{SOTA performance}. Comprehensive experiments on several widely adopted image SR datasets show that CubeFormer achieves superior detail recovery in HR images, consistently outperforming existing methods.
\end{itemize}

\begin{figure*}[t]
    \centering
    \includegraphics[width=\linewidth]{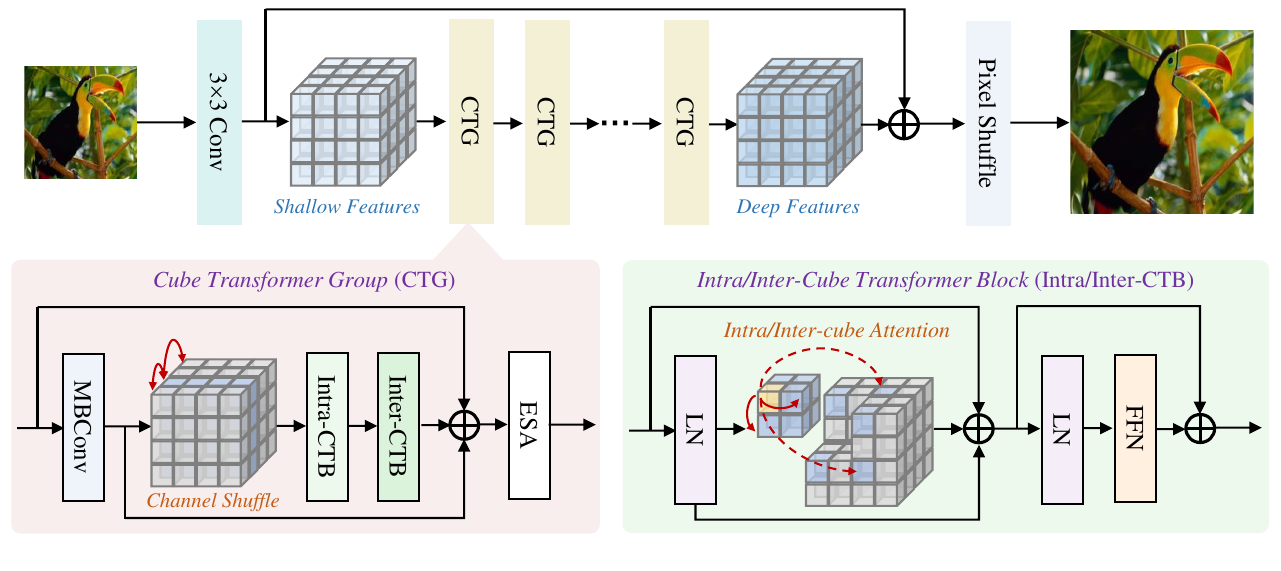}
    \vspace{-5mm}
    \caption{\textbf{Overall architecture of CubeFormer.} We utilize a convolutional layer to extract shallow features from the LR image and a series of Cube Transformer Groups (CTGs) for deep feature extraction. As shown in the bottom-left sub-figure, CTG incorporates the proposed Intra-CTB and Inter-CTB, channel shuffle, MBConv block, and enhanced spatial attention (ESA) block. In detail, Intra/Inter-CTB inherits the structure of the vanilla ViT and replaces self-attention with intra/inter cube attention as the bottom-right sub-figure illustrates.}
    \label{fig:ppl}
    \vspace{-3mm}
\end{figure*}
\section{Related Work}
\label{sec:related}

\noindent \textbf{CNN-based Image SR methods}. In the early years, CNN-based image SR models have achieved great success.
SRCNN \cite{dong2015image} is the first work to use deep neural network for image SR, which learns a map from a LR image to a corresponding HR image in an end-to-end manner.
DRCN \cite{hui2018fast} injected recursive operations and skip connections into the SR model to further degrade the model structure and achieve more efficient information flow.
CARN \cite{ahn2018fast} developed an end-to-end SR model with cascading residual networks. 
IMDN \cite{hui2019lightweight} introduced information multi-distillation block including distillation and selective fusion parts, where the distillation part focuses on capturing hierarchical information and the selective fusion part aggregated features at various scales.
HPUN \cite{sun2023hybrid} employed an efficient down-sampling module and enhance the feature representation and achieve better HR reconstruction.
In addition, some general strategies including knowledge distillation \cite{hinton2015distilling, gao2018image, zhang2021data} and neural architecture search \cite{chu2021fast} are also incorporated into SR model to further optimize model architecture and improve the efficiency of feature aggregation. 

\noindent \textbf{ViT-based Image SR methods}. 
More recently, researchers have developed many ViT-based solutions that show superior performance than CNN-based ones.
SwinIR \cite{liang2021swinir} introduced a strong baseline for image restoration, which incorporated swin transformer \cite{liu2021swin} layer and residual connection to promote feature exchange and high-quality image reconstruction.
ESRT \cite{lu2022transformer} combined CNN and transformer to construct a hybrid network that is composed of a lightweight CNN and transformer backbone. To be specific, these two backbones work collaboratively to extract deep features for better details recovery.
HAT \cite{chen2023activating} discovered that current SR methods only involve a limited spatial range of input information via attribution analysis. 
To alleviate this issue, HAT develops a new hybrid attention and an overlapping cross-attention module to fully explore the potential feature of the input image.
Moreover, some works \cite{chen2021pre, li2021efficient} attempted to introduce pre-training strategy to improve the learning effect of the networks on the SR task.

\noindent \textbf{Lightweight Vision Transformer.}
The attention mechanism was first proposed \cite{vaswani2017attention} and migrated to the field of computer vision as Vision Transformer (ViT) \cite{dosovitskiy2020image}, ViT-based networks have gradually replaced traditional CNN-based networks in many computer vision tasks for its outstanding performance.
Nevertheless, the vanilla ViT suffered from huge computational cost.
Hence, many attempts including \cite{mehta2021mobilevit,chen2022mobile,chen2022simple,yang2022lite,huang2022lightvit,li2023efficient} concentrated on further diminishing network parameters of ViT while preserving comparable results. 
LVT \cite{yang2022lite} delivered a lightweight ViT with two enhanced self-attention mechanisms: convolutional self-attention (CSA) and recursive atrous self-attention (RASA), which are used to extract low-level and high-level features, respectively.
MobileViT \cite{mehta2021mobilevit} offers a distinct approach to global modeling, presenting a vision backbone that integrates CNN and transformer architectures for enhanced representational capacity.
Besides, some methods \cite{wang2020linformer,li2023efficient} explored the matrix decomposition strategies, which can shrink the scale of matrix multiplication and further accelerate the inference.
\section{Method}
\label{sec:method}

%
\begin{figure*}[tp]
    \centering
    \includegraphics[width= 0.98 \textwidth]{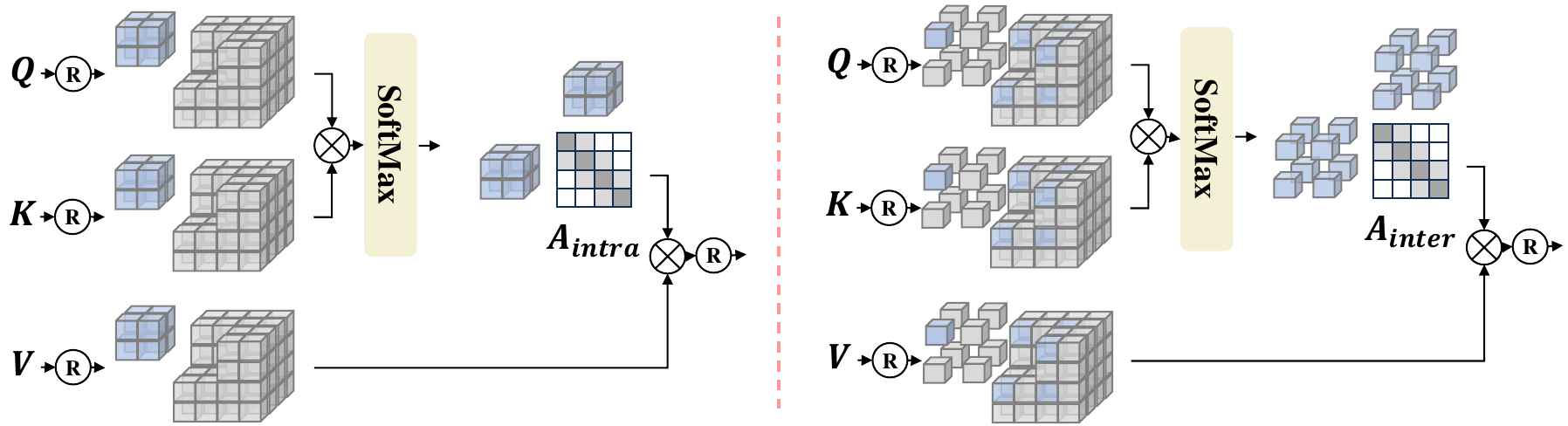}
     \vspace{-2mm}
     \begin{minipage}{.45\linewidth}
        \centering
        (a) Intra-cube Attention
    \end{minipage}
    \begin{minipage}{.45\linewidth}
        \centering
        (b) Inter-cube Attention
    \end{minipage}
    \caption{\textbf{Illustration of intra-cube and inter-cube attention.} The inputs are fed into convolutional layers to generate query, key, and value.
    Next, these components are further divided into 3D cubes and 3D cube grids by block and grid strategy, which are further used for obtaining affinity matrix.}
    \label{fig:cube_attn}
    \vspace{-3mm}
\end{figure*}


%
\subsection{CubeFormer}

\noindent \textbf{Overall pipeline}. 
The overall architecture of the proposed CubeFormer is illustrated in Figure \ref{fig:ppl}. 
Our CubeFormer accepts a LR degraded image as input and generates a HR image with more details.
There are three main processes in the pipeline, including feature extraction, backbone network, and HR image restoration.

\noindent \textbf{Feature extraction}. Given an input LR degraded image $I_{lr} \in \mathbb{R}^{H \times W \times 3}$, a $3 \times 3$ convolutional layer is firstly utilized to extract shallow features $X_{s} \in \mathbb{R}^{H \times W \times C}$ as: 
\begin{equation}
    X_{s} = H_{I}(I_{lr}), 
\end{equation}
where $H_{I}(\cdot)$ refers to the $3 \times 3$ convolution operation.

\noindent \textbf{Backbone network}. Next, a backbone network is used to extract deep features, which can be formulated as:
\begin{equation}
    X_{d} = H_{B}(X_{s}) + X_{s}, 
\end{equation}
where $X_{d}$ and $H_{B}(\cdot)$ denote the abstracted deep features and the transformation of backbone network. To be specific, the backbone network is composed of cascaded cube transformer groups (CTGs). In addition, a residual connection is injected to mine the potential of shallow features.

\noindent \textbf{HR image restoration}. In the end, a pixel shuffle-based \cite{shi2016real} HR image restoration module is applied to recover degraded details and reconstruct HR image as follows:
\begin{equation}
    I_{hr} = H_{H}(X_{d}), 
\end{equation}
where $H_{H}(\cdot)$ and $I_{hr}$ denote the transformation of HR image restoration module and the predicted HR image.

\noindent \textbf{Cube transformer group}.
The architecture of our cube transformer group (CTG) is shown in Figure \ref{fig:ppl}. We can see that the input sequentially passes through the MBConv block, channel shuffle, intra-cube attention block (Intra-CAB), inter-cube attention block (Inter-CAB), and the enhanced spatial attention (ESA) to obtain the output. MBConv block \cite{howard2017mobilenets, hu2018squeeze} and ESA \cite{liu2020residual} are used for pre-processing and post-processing. Channel shuffle operations \cite{zhang2018shufflenet} is introduced to promote information interaction in channel dimension. Inter-CAB and Intra-CAB are designed to complete global and local modeling, respectively.

\subsection{Cube Attention}

In prior research, homogeneous operators, such as spatial and channel attention, are commonly employed either individually or in a simple combination. 
However, these approaches may hinder the capacity to capture and infer nuanced low-level features. 
Hence, we develop cube attention, which generalizes 2D attention to 3D space to pursue exhaustive information interactions.
Leveraging Cube Attention, we further introduce two foundational modules, Inter-CAB and Intra-CAB, which incorporate block-based and grid-based sampling strategies, respectively.

\noindent \textbf{Cube attention}.
The Cube Attention paradigm is structured into four core stages: \textit{qkv} generation, cube embedding, affinity matrix calculation and cube merging.
Let the input 3D feature map be denoted as \( X_{in} \in \mathbb{R}^{H \times W \times C} \), where \( H \), \( W \), and \( C \) represent the spatial height, width, and channel dimensionality, respectively. Initially, \( X_{in} \) undergoes transformation via convolutional layers to derive the query \( Q \), key \( K \), and value \( V \), formally defined as:
\begin{equation}
\left\{
\begin{aligned}
    Q &= H_{Q}(X_{in}), \\
    K &= H_{K}(X_{in}), \\
    V &= H_{V}(X_{in}), \\
\end{aligned}
\right.
\end{equation}
where \( H_{Q} \), \( H_{K} \), and \( H_{V} \) denote the transformation functions responsible for the generation of query, key, and value, respectively.
Distinct from conventional window-based attention mechanisms, which partition the input into 2D panes, our cube attention mechanism segments the 3D space into non-overlapping cubes. 
Let \( Q_{cube} \in \mathbb{R}^{h \times w \times c} \), \( K_{cube} \in \mathbb{R}^{h \times w \times c} \), and \( V_{cube} \in \mathbb{R}^{h \times w \times c} \) denote individual cubes sampled from the query, key, and value representations, respectively. Each cube is then reshaped into a vector of dimension \( \mathbb{R}^{hwc} \).
Subsequently, an affinity matrix is computed by applying the softmax function to the inner product of the reshaped query and key vectors, yielding the output cube:
\begin{equation}
    X_{cube} = V_{cube}Softmax(Q_{cube}^TK_{cube}),
\end{equation}
where \( X_{cube} \) represents the resultant output cube, and \( Softmax(\cdot) \) denotes the softmax activation function. 
To ensure consistent spatial resolution, the output cubes are then merged, producing the final output \( X_{out} \in \mathbb{R}^{H \times W \times C} \).

\noindent \textbf{Intra-cube attention}.
Intra-cube attention implements self-attention across elements within each cube, enabling thorough local information interaction. 
The detailed mechanism of intra-cube attention is presented in Figure \ref{fig:cube_attn} (a).
During the cube partition phase, \( Q \), \( K \), and \( V \in \mathbb{R}^{H \times W \times C} \) are divided into non-overlapping 3D cubes of dimensions \( h_{1} \times w_{1} \times c_{1} \) via block sampling, formulated as follows:
\[
\begin{aligned}
    (H, W, C) & \to \left(\frac{H}{h_{1}} \times h_{1}, \frac{W}{w_{1}} \times w_{1}, \frac{C}{c_{1}} \times c_{1}\right) \\
    & \to \left(\frac{HWC}{h_{1}w_{1}c_{1}}, h_{1}w_{1}c_{1}\right).
\end{aligned}
\]
This sampling structure allows the affinity matrix to capture intricate voxel-wise relationships within each cube. 
In contrast to 2D window-based attention, which primarily models spatial relationships, the proposed intra-cube attention integrates a more extensive set of features by aggregating information within each local 3D cube, thus enhancing the feature representation's expressiveness.

\noindent \textbf{Inter-cube attention}.
Inter-cube attention enables cross-cube feature interactions, supporting comprehensive, long-range information abstraction. The architecture of inter-cube attention is detailed in Figure \ref{fig:cube_attn} (b).
During the cube partition stage, given \( Q \), \( K \), and \( V \in \mathbb{R}^{H \times W \times C} \), these tensors are partitioned into \( hwc \) 3D cube grids via grid sampling, where each cube has dimensions \( \frac{H}{h_{2}} \times \frac{W}{w_{2}} \times \frac{C}{c_{2}} \). This transformation can be formally described as:
\begin{equation}
    \begin{aligned}
        (H, W, C) & \to (h_{2} \times \frac{H}{h_{2}}, w_{2} \times \frac{W}{w_{2}}, c_{2} \times \frac{C}{c_{2}}) \\
        & \to (h_{2}w_{2}c_{2}, \frac{HWC}{h_{2}w_{2}c_{2}}). 
    \end{aligned}
\end{equation}
Inter-cube attention operates by modeling relationships between distinct cubes, functioning as a sparse global attention that efficiently aggregates information across extensive spatial extents. This design enhances the model's ability to capture cross-cube dependencies, fostering an omnidirectional feature abstraction that bolsters global perception.

\noindent \textbf{Intra/inter-cube transformer blocks}
We propose two specialized transformer blocks, intra-cube transformer block (Intra-CTB) and inter-cube transformer block (Inter-CTB), as the foundational components of our Cube Transformer Generator (CTG). 
Intra-CTB is designed to capture local details by leveraging intra-cube attention, enabling comprehensive information interaction within each individual cube. 
Conversely, Inter-CTB facilitates cross-cube information exchange, supporting the extraction of global sparse features for enhanced feature representation.
Both blocks adopt a two-stage architecture similar to the standard vision transformer but incorporate distinct attention mechanisms, as illustrated in Figure \ref{fig:ppl}. 

To begin with, the input features undergo layer normalization (LN) followed by intra- or inter-cube attention, which can be formally expressed as:
\begin{equation}
    F_{mid} = F_{in} + \text{Attn}(\text{LN}(F_{in})),
\end{equation}
where \( F_{in} \), \( F_{mid} \), \( \text{LN}(\cdot) \), and \( \text{Attn}(\cdot) \) represent the input features, intermediate features, layer normalization, and inter/intra-cube attention, respectively.

Next, \( F_{mid} \) is processed through an additional LN layer and a feed-forward network (FFN) layer, yielding the final output \( F_{out} \):
\begin{equation}
    F_{out} = F_{mid} + \text{FFN}(\text{LN}(F_{mid})),
\end{equation}
where \( \text{FFN}(\cdot) \) denotes the feed-forward network, comprising both convolution and depth-wise convolution layers.

\begin{figure}[t]
    \centering
    \includegraphics[width=1.0 \linewidth]{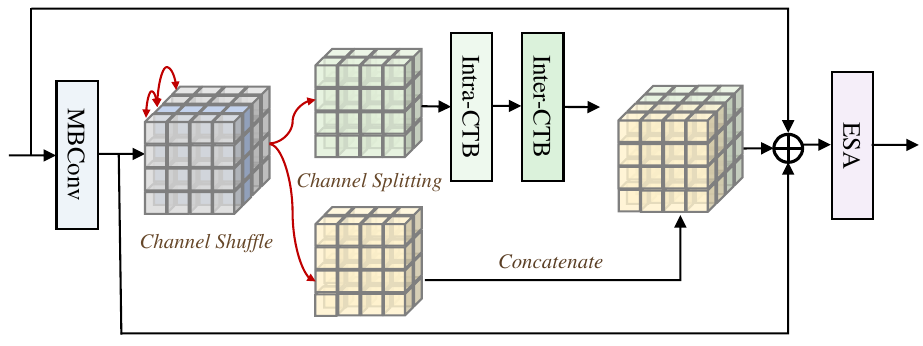}
    \caption{
    \textbf{Illustration of the lite version of Cube Transformer Group (CTG-lite).} In CTG-lite, an additional channel splitting operation is introduced following the channel shuffle, partitioning the features into two groups with half the original channel count.
    }
    \vspace{-3mm}
    \label{fig:ctg_lite}
\end{figure}

\subsection{CubeFormer-lite}
In this paper, we introduce CubeFormer-lite, a lite variant of the proposed CubeFormer tailored for efficient super-resolution. 
CubeFormer-lite achieves a substantial reduction in network parameters while preserving competitive performance. 
To construct CubeFormer-lite, we streamline the CTG in the original CubeFormer architecture, resulting in a simplified module referred to as CTG-lite.

\noindent \textbf{Channel Splitting.}  
As illustrated in Figure \ref{fig:ctg_lite}, CTG-lite introduces an additional channel splitting operation \cite{ma2018shufflenet} following the channel shuffle operation \cite{zhang2018shufflenet}, dividing the feature map into two equal groups along the channel dimension.
One group proceeds through the intra-CTB and inter-CTB modules for further processing, while the other group bypasses these computations and is later concatenated with the processed features. 
This selective feature involvement reduces the parameter count for intra-CTB and inter-CTB, thereby enhancing the efficiency of CTG-lite.

\noindent \textbf{CTG-lite}
Let the input feature to CTG-lite be \( X_{in} \in \mathbb{R}^{H \times W \times C} \). This input first undergoes processing through an MBConv block \cite{howard2017mobilenets, hu2018squeeze} followed by channel shuffle and channel splitting operations, resulting in a division of \( X_{in} \) into two equal parts: \( X_{1} \in \mathbb{R}^{H \times W \times \frac{C}{2}} \) and \( X_{2} \in \mathbb{R}^{H \times W \times \frac{C}{2}} \). This sequence of operations can be formally described as:
\noindent \begin{equation}
    X_{1}, X_{2} = \text{Split}(\text{Shuffle}(\text{MB}(X_{in}))),
\end{equation}
where \( \text{Split}(\cdot) \) and \( \text{Shuffle}(\cdot) \) represent the channel splitting and channel shuffle operations, respectively, and \( \text{MB}(\cdot) \) denotes the MBConv.

In this configuration, \( X_{1} \) proceeds through subsequent operations, while \( X_{2} \) bypasses these computations, later concatenated with the transformed \( \hat{X_{1}} \) to form \( \hat{X} \). The combined feature \( \hat{X} \) is then processed by the enhanced spatial attention (ESA) module \cite{liu2020residual} to produce the final output, \( X_{out} \).
The entire process can be expressed as follows:
\begin{equation}
\left\{
\begin{aligned}
    &\hat{X_{1}} = H_{inter}(H_{intra}(X_{1})), \\
    &\hat{X} = \text{Concat}(\hat{X_{1}}, X_{2}), \\
    &X_{out} = \text{ESA}(\hat{X}), \\
\end{aligned}
\right.
\end{equation}
where \( H_{intra}(\cdot) \) and \( H_{inter}(\cdot) \) correspond to the intra-CTB and inter-CTB transformations, \(\text{ESA}(\cdot) \) denotes the transformation within the ESA module, and \( \text{Concat}(\cdot, \cdot) \) represents the channel-wise concatenation operation.

\subsection{Learning Strategy}
To train our model effectively, spatial reconstruction loss and frequency reconstruction loss are employed.

\noindent \textbf{Spatial Reconstruction Loss.}
We begin by defining the spatial reconstruction loss, which encourages the predicted high-resolution image to approximate the ground truth image in the spatial domain. Specifically, the spatial reconstruction loss is formulated as:
\begin{equation}
    \mathcal{L}_{sr} = \Vert I_{hr} - I_{gt} \Vert_{1},
\end{equation}
where \( I_{gt} \) denotes the ground truth HR image, \( \mathcal{L}_{sr} \) is the reconstruction loss, and \( \Vert \cdot \Vert_{1} \) represents the \( l_1 \) norm.

\noindent \textbf{Frequency Reconstruction Loss.}
To further enhance CubeFormer’s ability to capture fine details, we introduce a frequency reconstruction loss \cite{cho2021rethinking, sun2022shufflemixer}, targeting high-frequency information. The frequency reconstruction loss is given by:
\begin{equation}
    \mathcal{L}_{fr} = \Vert \mathcal{F}(I_{hr}) - \mathcal{F}(I_{gt}) \Vert_{1}, 
\end{equation}
where \( \mathcal{L}_{fr} \) denotes the frequency reconstruction loss, and \( \mathcal{F}(\cdot) \) is the Fast Fourier Transform.

\noindent \textbf{Total Loss.}
The total objective function is then defined as a weighted combination of these two losses:
\begin{equation}
    \mathcal{L} = \mathcal{L}_{sr} + \lambda \mathcal{L}_{fr},
\end{equation}
where \( \mathcal{L} \) represents the total loss, and \( \lambda \) is a trade-off parameter set to 0.01 in our experiments.

\section{Experiments}
\label{sec:exp}

\begin{table}[t]
  \centering
  \scriptsize
  \setlength{\tabcolsep}{1.4mm}
    \caption{
    Quantitative comparison (PSNR/SSIM) of \textit{\textbf{lightweight image SR}} against state-of-the-art methods. From top to bottom, the results correspond to \(\times 2\), \(\times 3\), and \(\times 4\) super-resolution. Notably, CubeFormer consistently demonstrates superior performance across the B100, Urban100, and Manga109 datasets, establishing its efficacy in image quality enhancement.}
  \label{tab:psnr-ssim-lightweight}
	\begin{tabular}{l|c|c|c|c}
		\toprule[1pt]
		\multirow{2}{*}{Method} & Params & BSD100 & Urban100 & Manga109 \\
		& (K) & PSNR / SSIM & PSNR / SSIM & PSNR / SSIM \\
    \midrule
    \midrule
    IMDN~\cite{hui2019lightweight} & 694 & 32.19 / 0.8996 & 32.17 / 0.9283 & 38.88 / 0.9774 \\
    RFDN-L~\cite{liu2020residual} & 626 & 32.18 / 0.8996 & 32.24 / 0.9290  & 38.95 / 0.9773 \\
    MAFFSRN~\cite{muqeet2020multi} & 402 & 32.14 / 0.8994 & 31.96 / 0.9268 &  - / -\\
    LatticeNet~\cite{luo2020latticenet} & 756 & 32.25 / 0.9005 & 32.43 / 0.9302 & - / -\\
    SwinIR~\cite{liang2021swinir} & 878 & 32.31 / 0.9012 & 32.76 / 0.9340 & 39.12 / 0.9783 \\
    RLFN~\cite{kong2022residual} & 527 & 32.22 / 0.9000 & 32.33 / 0.9299 & - / -  \\
    GASSL-B+~\cite{wang2023global} & 689 & 32.28 / 0.9009 & 32.43 / 0.9310 & 39.10 / 0.9781 \\
    OSFFNet \cite{wang2024osffnet} & 516 & 32.29 / 0.9012 & 32.67 / 0.9331 & 39.09 / 0.9780 \\
    Omni-SR~\cite{wang2023omni} & 772 & 32.36 / 0.9020 & 33.05 / 0.9363  & 39.28 / 0.9784 \\
    SRConvNet-L \cite{li2024srconvnet} & 885 & 32.28 / 0.9010 & 32.59 / 0.9321 & 39.22 / 0.9779 \\
    \rowcolor{green!5}
    \textbf{CubeFormer} & 778 &  \textbf{32.36 / 0.9021} & \textbf{33.09 / 0.9368}  & \textbf{39.39 / 0.9786} \\
    \midrule
    \midrule
    IMDN~\cite{hui2019lightweight} & 703 & 29.09 / 0.8046 & 28.17 / 0.8519 & 33.61 / 0.9445 \\
    RFDN-L~\cite{liu2020residual} & 633 & 29.11 / 0.8053 & 28.32 / 0.8547  & 33.78 / 0.9458 \\
    MAFFSRN~\cite{muqeet2020multi} & 807 & 29.13 / 0.8061 & 28.26 / 0.8552 &  - / -\\
    LatticeNet~\cite{luo2020latticenet} & 765 & 29.15 / 0.8059 & 28.33 / 0.8538 & - / -\\
    SwinIR~\cite{liang2021swinir} & 886 & 29.20 / 0.8082& 28.66 / 0.8624& 33.98 / 0.9478 \\
    ESRT~\cite{lu2022transformer} & 770 & 29.15 / 0.8063 & 28.46 / 0.8574 & 33.95 / 0.9455 \\
    GASSL-B+~\cite{wang2023global} & 691 & 29.19 / 0.8070 & 28.39 / 0.8566 & 34.00 / 0.9467 \\
    OSFFNet~\cite{wang2024osffnet} & 524 & 29.21 / 0.8080 & 28.49 / 0.8595 & 34.00 / 0.9470 \\
    Omni-SR~\cite{wang2023omni} & 780 & 29.28 / 0.8094 & 28.84 / 0.8656  & 34.22 / 0.9487 \\
    SRConvNet-L \cite{li2024srconvnet} & 906 &  29.22 / 0.8081 & 28.56 / 0.8600 & 34.17 / 0.9479 \\
    \rowcolor{green!5}
    \textbf{CubeFormer} & 787 &  \textbf{29.29 / 0.8099} & \textbf{28.88 / 0.8664}  & \textbf{34.40 / 0.9490} \\
    \midrule
    \midrule
    IMDN~\cite{hui2019lightweight} & 715 & 27.56 / 0.7353 & 26.04 / 0.7838 & 30.45 / 0.9075 \\
    RFDN-L~\cite{liu2020residual} & 643 & 27.58 / 0.7363 & 26.20 / 0.7883 & 30.61 / 0.9096 \\
    MAFFSRN~\cite{muqeet2020multi} & 830 & 27.59 / 0.7370 & 26.16 / 0.7887 & - / - \\
    LatticeNet~\cite{luo2020latticenet} & 777 & 27.62 / 0.7367 & 26.25 / 0.7873 & - / -\\
    SwinIR~\cite{liang2021swinir} & 897 & 27.69 / 0.7406 & 26.47 / 0.7980 & 30.92 / 0.9151\\
    RLFN~\cite{kong2022residual} & 543 & 27.60 / 0.7364 & 26.17 / 0.7877 & - / -  \\
    ESRT~\cite{lu2022transformer} & 751 & 27.69 / 0.7379 & 26.39 / 0.7962 & 30.75 / 0.9100 \\
    GASSL-B+~\cite{wang2023global} & 694 & 27.66 / 0.7388 & 26.27 / 0.7914 & 30.92 / 0.9122 \\
    OSFFNet \cite{wang2024osffnet} & 537 &  27.66 / 0.7393 & 26.36 / 0.7950 & 30.84 / 0.9125 \\
    Omni-SR~\cite{wang2023omni} & 792 & 27.71 / 0.7415  & 26.64 / 0.8018 &  31.02 / 0.9151 \\
    SRConvNet-L \cite{li2024srconvnet} & 902 &   27.69 / 0.7402 & 26.47 / 0.7970 & 30.96 / 0.9139 \\
    \rowcolor{green!5}
    \textbf{CubeFormer} & 799 & \textbf{27.76 / 0.7422} & \textbf{26.66 / 0.8026}  & \textbf{31.39 / 0.9174} \\
    \bottomrule[1pt]
\end{tabular}
\vspace{-3mm}
\end{table}

\subsection{Experimental Setup}
\noindent \textbf{Datasets and Metrics.} Consistent with prior studies \cite{timofte2017ntire, wang2023omni}, we employ the DIV2K \cite{timofte2017ntire} dataset for training, evaluating model performance across the following established benchmarks: BSD100 \cite{martin2001database}, Urban100 \cite{huang2015single}, and Manga109 \cite{matsui2017sketch}.
We adopt PSNR and SSIM \cite{wang2004image} as evaluation metrics, both computed on the Y channel of the YCbCr space converted from RGB. Greater PSNR and SSIM values indicate superior super-resolution quality.

\noindent \textbf{Implementation Details.} In the training phase, LR images are synthesized by applying bicubic down-sampling \cite{zhang2017learning} to HR images at scaling factors of 2, 3, and 4.
Data augmentation techniques, including random horizontal flips, 90/270-degree rotations, and RGB channel shuffling, are employed to enhance model robustness.
Our CubeFormer utilize six CAG layers with 64 channels. For intra-cube and inter-cube attention, a four-head multi-head attention structure is applied. The intra-cube and inter-cube attention cubes are configured with dimensions \( h_{1} \times w_{1} \times c_{1} \) and grid sizes \( h_{2} \times w_{2} \times c_{2} \), both set to \( 8 \times 8 \times 4 \).
The training procedure employs the Adam optimizer \cite{kingma2014adam} with a batch size of 32 over 800K iterations. The initial learning rate of \( 5\times10^{-4} \) is halved every 200K iterations to ensure stable convergence.
For each batch, \( 64 \times 64 \) LR image patches are randomly cropped as inputs. All models are implemented in PyTorch and evaluated on an NVIDIA RTX 4090 GPU.

\begin{table}[t]
  \centering
  \scriptsize
  \setlength{\tabcolsep}{1.5mm}
    \caption{
    Quantitative comparison (PSNR/SSIM) for \textit{\textbf{efficient image SR}} with state-of-the-art methods. From top to bottom, the results correspond to \(\times 2\), \(\times 3\), and \(\times 4\) super-resolution. CubeFormer-lite attains superior PSNR and SSIM scores on the B100, Urban100, and Manga109 benchmarks, underscoring its effectiveness for image SR.}
  \label{tab:psnr-ssim-efficient}
	\begin{tabular}{l|c|c|c|c}
		\toprule[1pt]
		\multirow{2}{*}{Method} & Params & BSD100 & Urban100 & Manga109 \\
		& (K) & PSNR / SSIM & PSNR / SSIM & PSNR / SSIM \\
    \midrule
     \midrule
    RFDN~\cite{liu2020residual} & 534 & 32.16 / 0.8994 & 32.12 / 0.9278 & 38.88 / 0.9773 \\
    ShuffleMixer~\cite{sun2022shufflemixer} & 394 & 32.17 / 0.8995 & 31.89 / 0.9257 & 38.83 / 0.9774 \\
    RLFN-S~\cite{kong2022residual} & 454 & 32.19 / 0.8997 & 32.17 / 0.9286 & - / -  \\
    GASSL-S~\cite{wang2023global} & 280 & 32.14 / 0.8992 & 31.81 / 0.9253 & 38.57 / 0.9769 \\
    MLRN~\cite{gendy2023mixer}  & 488 & 32.21 / 0.9000 & 32.28 / 0.9297 & 38.76 / 0.9773 \\
    SAFMN~\cite{sun2023spatially}  & 228 & 32.16 / 0.8995 & 31.84 / 0.9256 & 38.71 / 0.9771 \\
    SeemoRe-T~\cite{zamfir2024see} & 220 & 32.23 / 0.9004 & 32.22 / 0.9286 & 39.01 / 0.9777 \\
    SRConvNet \cite{li2024srconvnet} & 387 &  32.16 / 0.8995 & 32.05 / 0.9272 & 38.87 / 0.9774 \\    
    \rowcolor{green!5}
    \textbf{CubeFormer-lite} & 349 &  \textbf{32.27 / 0.9011} & \textbf{32.60 / 0.9331}  & \textbf{39.04 / 0.9775} \\
    \midrule
    \midrule
    RFDN~\cite{liu2020residual} & 541 & 29.09 / 0.8050 & 28.21 / 0.8525 & 33.67 / 0.9449 \\
    ShuffleMixer~\cite{sun2022shufflemixer} & 415 & 29.12 / 0.8051 & 28.08 / 0.8498 & 33.69 / 0.9448 \\
    GASSL-S~\cite{wang2023global} & 373 & 29.06 / 0.8038 & 27.95 / 0.8474 & 33.42 / 0.9434 \\
    MLRN~\cite{gendy2023mixer}  & 496 & 29.10 / 0.8054 & 28.20 / 0.8533 & 33.66 / 0.9450 \\
    SAFMN~\cite{sun2023spatially}  & 233 & 29.08 / 0.8048 & 27.95 / 0.8474 & 33.52 / 0.9437 \\
    SeemoRe-T~\cite{zamfir2024see} & 225 & 29.15 / 0.8063 & 28.27 / 0.8538 & 33.92 / 0.9460 \\
    SRConvNet \cite{li2024srconvnet} & 387 & 29.08 / 0.8047 & 28.04 / 0.8498 & 33.51 / 0.9442 \\
    \rowcolor{green!5}
    \textbf{CubeFormer-lite} & 358 &  \textbf{29.20 / 0.8080} & \textbf{28.54 / 0.8601}  & \textbf{34.04 / 0.9471} \\
    \midrule
    \midrule
    RFDN~\cite{liu2020residual} & 550 & 27.57 / 0.7360 & 26.11 / 0.7858 & 30.58 / 0.9089 \\
    ShuffleMixer~\cite{sun2022shufflemixer} & 411 & 27.61 / 0.7366 & 26.08 / 0.7835 & 30.65 / 0.9093 \\
    RLFN-S~\cite{kong2022residual} & 470 & 27.58 / 0.7359 & 26.15 / 0.7866 & - / -  \\
    GASSL-S~\cite{wang2023global} & 428 & 27.56 / 0.7351 & 25.98 / 0.7818 & 30.35 / 0.9070 \\
    MLRN~\cite{gendy2023mixer} & 507 & 27.57 / 0.7365 & 26.10 / 0.7867 & 30.56 / 0.9092 \\
    SAFMN~\cite{sun2023spatially}  & 240 & 27.58 / 0.7359 & 25.97 / 0.7809 & 30.43 / 0.9063 \\
    SeemoRe-T~\cite{zamfir2024see} & 232 & 27.65 / 0.7384 & 26.23 / 0.7883 & 30.82 / 0.9107 \\
    SRConvNet \cite{li2024srconvnet} & 382 & 27.57 / 0.7359 & 26.06 / 0.7845 & 30.35 / 0.9075 \\
    \rowcolor{green!5}
    \textbf{CubeFormer-lite} & 370 &  \textbf{27.67 / 0.7392} & \textbf{26.37 / 0.7939}  & \textbf{30.94 / 0.9129} \\
\bottomrule[1pt]
\end{tabular}
\vspace{-3mm}
\end{table}

\subsection{Comparison with the State-of-the-art Methods}
\noindent \textbf{Compared methods}. We conduct the comparisons on both lightweight and efficient SR. To be specific, the compared solutions contain IMDN \cite{hui2019lightweight}, RFDN \cite{liu2020residual}, LatticeNet \cite{luo2020latticenet}, SwinIR \cite{liang2021swinir}, RLFN \cite{kong2022residual}, ESRT \cite{lu2022transformer}, Shufflemixer \cite{sun2022shufflemixer}, GASSL \cite{wang2023global}, MLRN \cite{gendy2023mixer}, SAFMN \cite{sun2023spatially}, OSFFNet~\cite{wang2024osffnet}, SeemoRe~\cite{zamfir2024see}, SRConvNet \cite{li2024srconvnet} and OmniSR \cite{wang2023omni}.

\noindent \textbf{Quantitative results}. 
The comparison results for lightweight SR across benchmark datasets are presented in Table \ref{tab:psnr-ssim-lightweight}, revealing several key insights.
In general, CubeFormer achieves either the best performance across all scale factors. 
Additionally, heterogeneous structure-based methods, including OmniSR \cite{wang2023omni} and CubeFormer, consistently outperform homogeneous approaches like SwinIR, underscoring the effectiveness of heterogeneous and omnidirectional information abstraction.
The numerical results for efficient SR are summarized in Table \ref{tab:psnr-ssim-efficient}. 
CubeFormer-lite excels on the BSD100, Urban100, and Manga109 benchmarks, setting new performance standards. 
Compared with state-of-the-art efficient SR methods, such as ShuffleMixer, GASSL-S, MLRN, and SAFMN, CubeFormer-lite achieves superior results while maintaining a lower parameter count, highlighting its efficiency and capability in resource-constrained environments.

\begin{figure*}
    \centering
    \includegraphics[width=\linewidth]{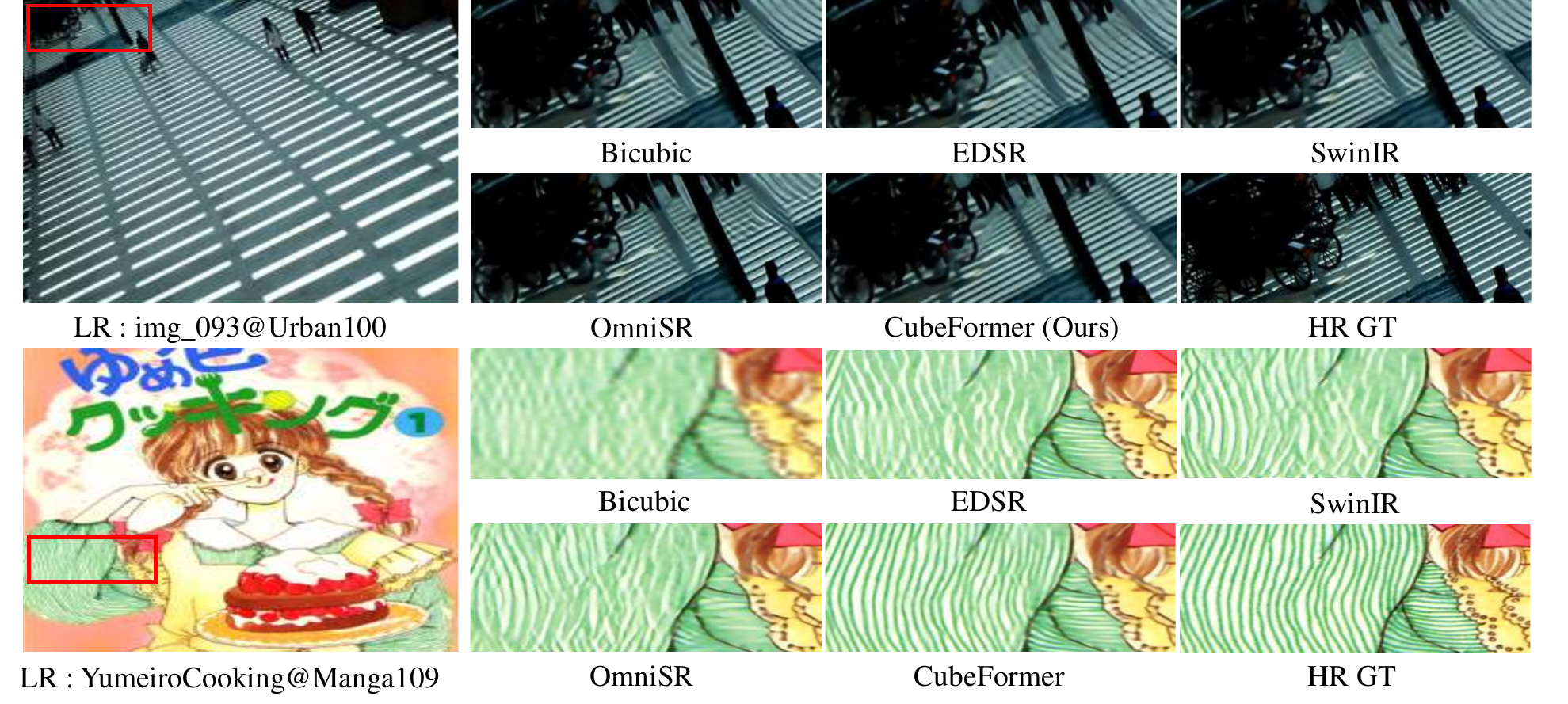}
    \vspace{-7mm}
    \caption{\textbf{Visual comparison of our CubeFormer and other state-of-the-art lightweight SR methods including bicubic interpolation, classical CNN-based method EDSR \cite{lim2017enhanced}, state-of-the-art methods SwinIR \cite{liang2021swinir} and OmniSR \cite{wang2023omni}.} Tow images from Urban100 dataset \cite{huang2015single} and Manga109 dataset \cite{matsui2017sketch} are picked for visualization. Our CubeFormer recovers more accurate details while other methods may have some drawbacks including blurring or artifacts.}
    \label{fig:vis}
    \vspace{-5mm}
\end{figure*}

\noindent \textbf{Qualitative Results.} To further demonstrate the effectiveness of the proposed CubeFormer, we present visual comparisons against SoTA methods, including EDSR, SwinIR, and OmniSR. \
Figure \ref{fig:vis} illustrates these comparisons on the Urban100 and Manga109 datasets.
In the first image, close examination of enlarged local patches reveals that alternative methods tend to produce blurred edges or artifacts. 
In contrast, CubeFormer successfully restores sharper, more refined details, closely approximating the ground truth HR image.
For the second image from the Manga109 dataset, methods such as EDSR, SwinIR, and OmniSR yield unclear green lines, while CubeFormer reconstructs more intricate textures with improved clarity.
Overall, CubeFormer demonstrates an enhanced capability in capturing fine-grained low-level features, effectively restoring missing local details in degraded low-resolution images.

\begin{figure}[t]
    \centering
    \includegraphics[width= 1.0 \linewidth]{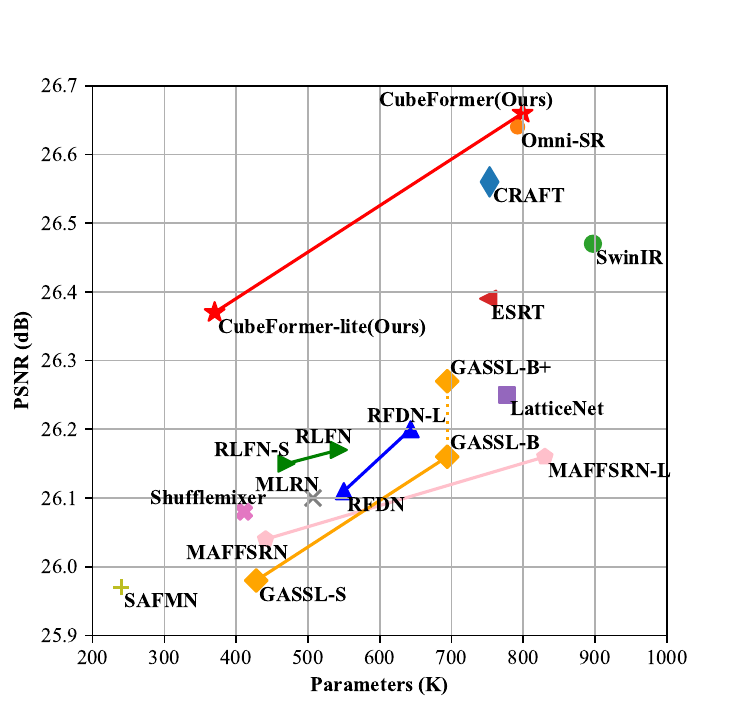}
    \caption{\textbf{PSNR vs. the number of model parameters of different lightweight SR methods on Urban100 dataset under x4 scale.} Both the proposed CubeFormer and CubeFormer-lite achieve the best performance at their respective parameter scales.}
    \label{fig:res}
    \vspace{-6mm}
\end{figure}

%

\noindent \textbf{Parameter Analysis.}
We present a PSNR-parameter plot comparing state-of-the-art lightweight and efficient SR methods on the Urban100 dataset at a \(\times 4\) scaling factor, as shown in Figure \ref{fig:res}.
The results reveal that CubeFormer consistently surpasses other methods across varying model sizes. Notably, as parameter count increases, CubeFormer demonstrates a more favorable PSNR improvement trend compared to its counterparts, while some competing methods exhibit diminishing returns in PSNR despite additional parameters.
This analysis underscores CubeFormer’s strength in achieving an optimal balance between model complexity and performance, establishing it as a superior solution for lightweight and efficient SR tasks.

\subsection{Ablation Study}


\noindent \textbf{Effectiveness of Cube Attention.}  
We begin by analyzing the impact of cube attention. To this end, three additional sets of experiments were conducted where the cube attention modules in CTG were substituted with spatial attention, channel attention, and a combination of both. The spatial attention is derived from SwinIR \cite{liang2021swinir}, while the channel attention originates from Restormer \cite{zamir2022restormer}.
The results demonstrate that models using homogeneous structures, such as those with only \textbf{SA} or only \textbf{CA}, show relatively inferior performance compared to heterogeneous structures, like the combination of SA and CA or our proposed cube attention. 
This outcome highlights the benefits of heterogeneous modeling for enhanced detail recovery in high-resolution images.
Notably, CubeFormer surpasses even the combined \textbf{SA + CA} configuration, indicating that the proposed cube attention achieves more effective detail recovery than the synchronous spatial and channel attention in \textbf{SA + CA}.

\begin{table}[t]
    \centering
    \small
    \setlength{\tabcolsep}{2.2mm}
    \caption{
    \textbf{Ablation study of Cube Attention on Urban100 and Manga109 datasets under x2 scale.} To rigorously assess the effectiveness of the proposed cube attention mechanism, we conduct comparative experiments where CubeFormer is modified by replacing cube attention with spatial attention (SA), channel attention (CA), and a combined synchronous attention (SA + CA).}
    \vspace{-1mm}
    \label{tab:ablation1}
	\begin{tabular}{l|cc|cc}
		\toprule[1pt]
		\multirow{2}{*}{Model} & \multicolumn{2}{c|}{Urban100} & \multicolumn{2}{c}{Manga109} \\
            & PSNR & SSIM & PSNR & SSIM\\
            \midrule
            \midrule
            \textbf{SA} & 32.84&0.9350 & 39.31&0.9783\\
            \textbf{CA} & 32.76&0.9344 & 39.29&0.9783\\
            \textbf{SA+CA} & 33.00&0.9364 & 39.33&0.9783\\
            \midrule
            \textbf{CubeFormer} & 33.09&0.9368 & 39.39&0.9786\\
        \bottomrule[1pt]
    \end{tabular}
    
\vspace{-3mm}
\end{table}

\noindent \textbf{Effectiveness of intra/Inter-CTB}.
We further evaluate the impact of Intra/Inter-CTB by performing four extra experiments.
Table \ref{tab:ablation2} reports the numerical performance on Urban100 dataset under $\times$2 scale. As shown in Table \ref{tab:ablation2}, we list four models to be compared. \textbf{Intra-CTB} and \textbf{Inter-CTB} represent removing the Inter-CTB or Intra-CTB from CubeFormer, respectively. \textbf{Inter-CTB-2} and \textbf{Intra-CTB-2} denote that Inter-CTB and Intra-CTB replace each other. We see that there are performance degradation when any of Inter-CTB and Intra-CTB is absent. This is attributed to the capability of Intra-CTB for extracting local features and the role of Inter-CTB in facilitating comprehensive global information extraction. These functionalities collectively contribute to enhancing the quality of restored HR images.

\begin{table}[t]
    \centering
    \small
    \setlength{\tabcolsep}{1.2mm}
    \caption{
    \textbf{Ablation study of Intra/Inter-CTB on Urban100 dataset under x2 scale.} Our CubeFormer, employing the concurrent utilization of Intra-CTB and Inter-CTB, achieves the best performance. The outcomes show the effectiveness of the global and local modeling capabilities of Intra-CTB and Inter-CTB.}
    \label{tab:ablation2}
    \vspace{-1mm}

	\begin{tabular}{l|cc|cc|cc}
		\toprule[1pt]
            \multirow{2}{*}{Model} & \multicolumn{2}{c|}{Block 1} & \multicolumn{2}{c|}{Block 2} & \multirow{2}{*}{PSNR} & \multirow{2}{*}{SSIM} \\
            & Intra & Inter & Intra & Inter & \\
            \midrule
            \midrule
            \textbf{Intra-CTB} & \checkmark & & & & 32.64 & 0.9332\\
            \textbf{Inter-CTB} & & \checkmark & & & 32.63 & 0.9336\\
            \textbf{Intra-CTB-2} & \checkmark & & \checkmark & & 32.86 & 0.9352\\
            \textbf{Inter-CTB-2} & & \checkmark & & \checkmark & 32.87 & 0.9354\\
            \midrule
            \textbf{CubeFormer} & \checkmark & & & \checkmark & 33.09 & 0.9368\\
        \bottomrule[1pt]
    \end{tabular}
    
\vspace{-3mm}
\end{table}
\section{Conclusion}
\label{sec:conclusion}

In this study, we have explored the limitations in current ViT-based methods, where there is a deficiency in feature richness, leading to an inability to extract diverse low-level features. 
This shortfall results in unclear textures and artifacts in the restored HR images. 
We therefore propose CubeFormer, a novel lightweight SR framework designed to capture more finer-grained details.
Specifically, we introduce cube attention, a novel mechanism that extends 2D attention to the 3D space. 
This innovation enables CubeFormer to perform exhaustive information interactions and abstract more intricate features. Additionally, we introduce two transformer blocks, namely Intra-Cube Transformer Block (Intra-CTB) and Inter-Cube Transformer Block (Inter-CTB), to concurrently model global and local context.
The experiments on widely-used datasets show that CubeFormer achieves SoTA SR performance and obtains better visual results with more details.

{
    \small
    \bibliographystyle{ieeenat_fullname}
    \bibliography{main}

\begin{thebibliography}{53}
\providecommand{\natexlab}[1]{#1}
\providecommand{\url}[1]{\texttt{#1}}
\expandafter\ifx\csname urlstyle\endcsname\relax
  \providecommand{\doi}[1]{doi: #1}\else
  \providecommand{\doi}{doi: \begingroup \urlstyle{rm}\Url}\fi

\bibitem[Ahn et~al.(2018)Ahn, Kang, and Sohn]{ahn2018fast}
Namhyuk Ahn, Byungkon Kang, and Kyung-Ah Sohn.
\newblock Fast, accurate, and lightweight super-resolution with cascading residual network.
\newblock In \emph{Proceedings of the European Conference on Computer Vision}, 2018.

\bibitem[Chen et~al.(2021)Chen, Wang, Guo, Xu, Deng, Liu, Ma, Xu, Xu, and Gao]{chen2021pre}
Hanting Chen, Yunhe Wang, Tianyu Guo, Chang Xu, Yiping Deng, Zhenhua Liu, Siwei Ma, Chunjing Xu, Chao Xu, and Wen Gao.
\newblock Pre-trained image processing transformer.
\newblock In \emph{Proceedings of the IEEE/CVF Conference on Computer Vision and Pattern Recognition}, 2021.

\bibitem[Chen et~al.(2022{\natexlab{a}})Chen, Chu, Zhang, and Sun]{chen2022simple}
Liangyu Chen, Xiaojie Chu, Xiangyu Zhang, and Jian Sun.
\newblock Simple baselines for image restoration.
\newblock In \emph{European Conference on Computer Vision}, 2022{\natexlab{a}}.

\bibitem[Chen et~al.(2023)Chen, Wang, Zhou, Qiao, and Dong]{chen2023activating}
Xiangyu Chen, Xintao Wang, Jiantao Zhou, Yu Qiao, and Chao Dong.
\newblock Activating more pixels in image super-resolution transformer.
\newblock In \emph{Proceedings of the IEEE/CVF Conference on Computer Vision and Pattern Recognition}, 2023.

\bibitem[Chen et~al.(2022{\natexlab{b}})Chen, Dai, Chen, Liu, Dong, Yuan, and Liu]{chen2022mobile}
Yinpeng Chen, Xiyang Dai, Dongdong Chen, Mengchen Liu, Xiaoyi Dong, Lu Yuan, and Zicheng Liu.
\newblock Mobile-former: Bridging mobilenet and transformer.
\newblock In \emph{Proceedings of the IEEE/CVF Conference on Computer Vision and Pattern Recognition}, 2022{\natexlab{b}}.

\bibitem[Cho et~al.(2021)Cho, Ji, Hong, Jung, and Ko]{cho2021rethinking}
Sung-Jin Cho, Seo-Won Ji, Jun-Pyo Hong, Seung-Won Jung, and Sung-Jea Ko.
\newblock Rethinking coarse-to-fine approach in single image deblurring.
\newblock In \emph{Proceedings of the IEEE/CVF International Conference on Computer Vision}, 2021.

\bibitem[Chu et~al.(2021)Chu, Zhang, Ma, Xu, and Li]{chu2021fast}
Xiangxiang Chu, Bo Zhang, Hailong Ma, Ruijun Xu, and Qingyuan Li.
\newblock Fast, accurate and lightweight super-resolution with neural architecture search.
\newblock In \emph{2020 25th International Conference on Pattern Recognition}, 2021.

\bibitem[Dong et~al.(2015)Dong, Loy, He, and Tang]{dong2015image}
Chao Dong, Chen~Change Loy, Kaiming He, and Xiaoou Tang.
\newblock Image super-resolution using deep convolutional networks.
\newblock \emph{IEEE Transactions on Pattern Analysis and Machine Intelligence}, 2015.

\bibitem[Dosovitskiy et~al.(2020)Dosovitskiy, Beyer, Kolesnikov, Weissenborn, Zhai, Unterthiner, Dehghani, Minderer, Heigold, Gelly, et~al.]{dosovitskiy2020image}
Alexey Dosovitskiy, Lucas Beyer, Alexander Kolesnikov, Dirk Weissenborn, Xiaohua Zhai, Thomas Unterthiner, Mostafa Dehghani, Matthias Minderer, Georg Heigold, Sylvain Gelly, et~al.
\newblock An image is worth 16x16 words: Transformers for image recognition at scale.
\newblock \emph{arXiv preprint arXiv:2010.11929}, 2020.

\bibitem[Gao et~al.(2018)Gao, Zhao, Li, and Tong]{gao2018image}
Qinquan Gao, Yan Zhao, Gen Li, and Tong Tong.
\newblock Image super-resolution using knowledge distillation.
\newblock In \emph{Asian Conference on Computer Vision}, 2018.

\bibitem[Gendy et~al.(2023)Gendy, Sabor, Hou, and He]{gendy2023mixer}
Garas Gendy, Nabil Sabor, Jingchao Hou, and Guanghui He.
\newblock Mixer-based local residual network for lightweight image super-resolution.
\newblock In \emph{Proceedings of the IEEE/CVF Conference on Computer Vision and Pattern Recognition}, 2023.

\bibitem[Hinton et~al.(2015)Hinton, Vinyals, and Dean]{hinton2015distilling}
Geoffrey Hinton, Oriol Vinyals, and Jeff Dean.
\newblock Distilling the knowledge in a neural network.
\newblock \emph{arXiv preprint arXiv:1503.02531}, 2015.

\bibitem[Howard et~al.(2017)Howard, Zhu, Chen, Kalenichenko, Wang, Weyand, Andreetto, and Adam]{howard2017mobilenets}
Andrew~G Howard, Menglong Zhu, Bo Chen, Dmitry Kalenichenko, Weijun Wang, Tobias Weyand, Marco Andreetto, and Hartwig Adam.
\newblock Mobilenets: Efficient convolutional neural networks for mobile vision applications.
\newblock \emph{arXiv preprint arXiv:1704.04861}, 2017.

\bibitem[Hu et~al.(2018)Hu, Shen, and Sun]{hu2018squeeze}
Jie Hu, Li Shen, and Gang Sun.
\newblock Squeeze-and-excitation networks.
\newblock In \emph{Proceedings of the IEEE Conference on Computer Vision and Pattern Recognition}, 2018.

\bibitem[Huang et~al.(2015)Huang, Singh, and Ahuja]{huang2015single}
Jia-Bin Huang, Abhishek Singh, and Narendra Ahuja.
\newblock Single image super-resolution from transformed self-exemplars.
\newblock In \emph{Proceedings of the IEEE Conference on Computer Vision and Pattern Recognition}, 2015.

\bibitem[Huang et~al.(2022)Huang, Huang, You, Wang, Qian, and Xu]{huang2022lightvit}
Tao Huang, Lang Huang, Shan You, Fei Wang, Chen Qian, and Chang Xu.
\newblock Lightvit: Towards light-weight convolution-free vision transformers.
\newblock \emph{arXiv preprint arXiv:2207.05557}, 2022.

\bibitem[Hui et~al.(2018)Hui, Wang, and Gao]{hui2018fast}
Zheng Hui, Xiumei Wang, and Xinbo Gao.
\newblock Fast and accurate single image super-resolution via information distillation network.
\newblock In \emph{Proceedings of the IEEE Conference on Computer Vision and Pattern Recognition}, 2018.

\bibitem[Hui et~al.(2019)Hui, Gao, Yang, and Wang]{hui2019lightweight}
Zheng Hui, Xinbo Gao, Yunchu Yang, and Xiumei Wang.
\newblock Lightweight image super-resolution with information multi-distillation network.
\newblock In \emph{Proceedings of the 27th ACM International Conference on Multimedia}, 2019.

\bibitem[Kim et~al.(2016)Kim, Lee, and Lee]{kim2016accurate}
Jiwon Kim, Jung~Kwon Lee, and Kyoung~Mu Lee.
\newblock Accurate image super-resolution using very deep convolutional networks.
\newblock In \emph{Proceedings of the IEEE Conference on Computer Vision and Pattern Recognition}, 2016.

\bibitem[Kingma and Ba(2014)]{kingma2014adam}
Diederik~P Kingma and Jimmy Ba.
\newblock Adam: A method for stochastic optimization.
\newblock \emph{arXiv preprint arXiv:1412.6980}, 2014.

\bibitem[Kong et~al.(2022)Kong, Li, Liu, Liu, He, Bai, Chen, and Fu]{kong2022residual}
Fangyuan Kong, Mingxi Li, Songwei Liu, Ding Liu, Jingwen He, Yang Bai, Fangmin Chen, and Lean Fu.
\newblock Residual local feature network for efficient super-resolution.
\newblock In \emph{Proceedings of the IEEE/CVF Conference on Computer Vision and Pattern Recognition}, 2022.

\bibitem[Li et~al.(2024)Li, Cong, Wu, Bai, Wang, and Zhao]{li2024srconvnet}
Feng Li, Runmin Cong, Jingjing Wu, Huihui Bai, Meng Wang, and Yao Zhao.
\newblock Srconvnet: A transformer-style convnet for lightweight image super-resolution.
\newblock \emph{International Journal of Computer Vision}, 2024.

\bibitem[Li et~al.(2021)Li, Lu, Qian, Lu, Zhang, and Jia]{li2021efficient}
Wenbo Li, Xin Lu, Shengju Qian, Jiangbo Lu, Xiangyu Zhang, and Jiaya Jia.
\newblock On efficient transformer-based image pre-training for low-level vision.
\newblock \emph{arXiv preprint arXiv:2112.10175}, 2021.

\bibitem[Li et~al.(2023)Li, Fan, Xiang, Demandolx, Ranjan, Timofte, and Van~Gool]{li2023efficient}
Yawei Li, Yuchen Fan, Xiaoyu Xiang, Denis Demandolx, Rakesh Ranjan, Radu Timofte, and Luc Van~Gool.
\newblock Efficient and explicit modelling of image hierarchies for image restoration.
\newblock In \emph{Proceedings of the IEEE/CVF Conference on Computer Vision and Pattern Recognition}, 2023.

\bibitem[Liang et~al.(2021)Liang, Cao, Sun, Zhang, Van~Gool, and Timofte]{liang2021swinir}
Jingyun Liang, Jiezhang Cao, Guolei Sun, Kai Zhang, Luc Van~Gool, and Radu Timofte.
\newblock Swinir: Image restoration using swin transformer.
\newblock In \emph{Proceedings of the IEEE/CVF International Conference on Computer Vision}, 2021.

\bibitem[Lim et~al.(2017)Lim, Son, Kim, Nah, and Mu~Lee]{lim2017enhanced}
Bee Lim, Sanghyun Son, Heewon Kim, Seungjun Nah, and Kyoung Mu~Lee.
\newblock Enhanced deep residual networks for single image super-resolution.
\newblock In \emph{Proceedings of the IEEE Conference on cComputer Vision and Pattern Recognition workshops}, 2017.

\bibitem[Liu et~al.(2020)Liu, Tang, and Wu]{liu2020residual}
Jie Liu, Jie Tang, and Gangshan Wu.
\newblock Residual feature distillation network for lightweight image super-resolution.
\newblock In \emph{Proceedings of the European Conference on Computer Vision}, 2020.

\bibitem[Liu et~al.(2021)Liu, Lin, Cao, Hu, Wei, Zhang, Lin, and Guo]{liu2021swin}
Ze Liu, Yutong Lin, Yue Cao, Han Hu, Yixuan Wei, Zheng Zhang, Stephen Lin, and Baining Guo.
\newblock Swin transformer: Hierarchical vision transformer using shifted windows.
\newblock In \emph{Proceedings of the IEEE/CVF International Conference on Computer Vision}, 2021.

\bibitem[Lu et~al.(2022)Lu, Li, Liu, Huang, Zhang, and Zeng]{lu2022transformer}
Zhisheng Lu, Juncheng Li, Hong Liu, Chaoyan Huang, Linlin Zhang, and Tieyong Zeng.
\newblock Transformer for single image super-resolution.
\newblock In \emph{Proceedings of the IEEE/CVF Conference on Computer Vision and Pattern Recognition}, 2022.

\bibitem[Luo et~al.(2020)Luo, Xie, Zhang, Qu, Li, and Fu]{luo2020latticenet}
Xiaotong Luo, Yuan Xie, Yulun Zhang, Yanyun Qu, Cuihua Li, and Yun Fu.
\newblock Latticenet: Towards lightweight image super-resolution with lattice block.
\newblock In \emph{Proceedings of the European Conference on Computer Vision}, 2020.

\bibitem[Ma et~al.(2018)Ma, Zhang, Zheng, and Sun]{ma2018shufflenet}
Ningning Ma, Xiangyu Zhang, Hai-Tao Zheng, and Jian Sun.
\newblock Shufflenet v2: Practical guidelines for efficient cnn architecture design.
\newblock In \emph{Proceedings of the European Conference on Computer Vision}, 2018.

\bibitem[Martin et~al.(2001)Martin, Fowlkes, Tal, and Malik]{martin2001database}
David Martin, Charless Fowlkes, Doron Tal, and Jitendra Malik.
\newblock A database of human segmented natural images and its application to evaluating segmentation algorithms and measuring ecological statistics.
\newblock In \emph{Proceedings of IEEE International Conference on Computer Vision}, 2001.

\bibitem[Matsui et~al.(2017)Matsui, Ito, Aramaki, Fujimoto, Ogawa, Yamasaki, and Aizawa]{matsui2017sketch}
Yusuke Matsui, Kota Ito, Yuji Aramaki, Azuma Fujimoto, Toru Ogawa, Toshihiko Yamasaki, and Kiyoharu Aizawa.
\newblock Sketch-based manga retrieval using manga109 dataset.
\newblock \emph{Multimedia Tools and Applications}, 2017.

\bibitem[Mehta and Rastegari(2021)]{mehta2021mobilevit}
Sachin Mehta and Mohammad Rastegari.
\newblock Mobilevit: light-weight, general-purpose, and mobile-friendly vision transformer.
\newblock \emph{arXiv preprint arXiv:2110.02178}, 2021.

\bibitem[Muqeet et~al.(2020)Muqeet, Hwang, Yang, Kang, Kim, and Bae]{muqeet2020multi}
Abdul Muqeet, Jiwon Hwang, Subin Yang, JungHeum Kang, Yongwoo Kim, and Sung-Ho Bae.
\newblock Multi-attention based ultra lightweight image super-resolution.
\newblock In \emph{Proceedings of the European Conference on Computer Vision}, 2020.

\bibitem[Shi et~al.(2016)Shi, Caballero, Husz{\'a}r, Totz, Aitken, Bishop, Rueckert, and Wang]{shi2016real}
Wenzhe Shi, Jose Caballero, Ferenc Husz{\'a}r, Johannes Totz, Andrew~P Aitken, Rob Bishop, Daniel Rueckert, and Zehan Wang.
\newblock Real-time single image and video super-resolution using an efficient sub-pixel convolutional neural network.
\newblock In \emph{Proceedings of the IEEE Conference on Computer Vision and Pattern Recognition}, 2016.

\bibitem[Sun et~al.(2023{\natexlab{a}})Sun, Zhang, Jiang, and Fu]{sun2023hybrid}
Bin Sun, Yulun Zhang, Songyao Jiang, and Yun Fu.
\newblock Hybrid pixel-unshuffled network for lightweight image super-resolution.
\newblock In \emph{Proceedings of the AAAI Conference on Artificial Intelligence}, 2023{\natexlab{a}}.

\bibitem[Sun et~al.(2022)Sun, Pan, and Tang]{sun2022shufflemixer}
Long Sun, Jinshan Pan, and Jinhui Tang.
\newblock Shufflemixer: An efficient convnet for image super-resolution.
\newblock \emph{Advances in Neural Information Processing Systems}, 2022.

\bibitem[Sun et~al.(2023{\natexlab{b}})Sun, Dong, Tang, and Pan]{sun2023spatially}
Long Sun, Jiangxin Dong, Jinhui Tang, and Jinshan Pan.
\newblock Spatially-adaptive feature modulation for efficient image super-resolution.
\newblock \emph{Proceedings of the IEEE International Conference on Computer Vision}, 2023{\natexlab{b}}.

\bibitem[Timofte et~al.(2017)Timofte, Agustsson, Van~Gool, Yang, and Zhang]{timofte2017ntire}
Radu Timofte, Eirikur Agustsson, Luc Van~Gool, Ming-Hsuan Yang, and Lei Zhang.
\newblock Ntire 2017 challenge on single image super-resolution: Methods and results.
\newblock In \emph{Proceedings of the IEEE Conference on Computer Vision and Pattern Recognition workshops}, 2017.

\bibitem[Vaswani et~al.(2017)Vaswani, Shazeer, Parmar, Uszkoreit, Jones, Gomez, Kaiser, and Polosukhin]{vaswani2017attention}
Ashish Vaswani, Noam Shazeer, Niki Parmar, Jakob Uszkoreit, Llion Jones, Aidan~N Gomez, {\L}ukasz Kaiser, and Illia Polosukhin.
\newblock Attention is all you need.
\newblock \emph{Advances in Neural Information Processing Systems}, 2017.

\bibitem[Wang et~al.(2023{\natexlab{a}})Wang, Chen, Ni, Liu, and Liu]{wang2023omni}
Hang Wang, Xuanhong Chen, Bingbing Ni, Yutian Liu, and Jinfan Liu.
\newblock Omni aggregation networks for lightweight image super-resolution.
\newblock In \emph{Proceedings of the IEEE/CVF Conference on Computer Vision and Pattern Recognition}, 2023{\natexlab{a}}.

\bibitem[Wang et~al.(2023{\natexlab{b}})Wang, Zhang, Qin, Van~Gool, and Fu]{wang2023global}
Huan Wang, Yulun Zhang, Can Qin, Luc Van~Gool, and Yun Fu.
\newblock Global aligned structured sparsity learning for efficient image super-resolution.
\newblock \emph{IEEE Transactions on Pattern Analysis and Machine Intelligence}, 2023{\natexlab{b}}.

\bibitem[Wang et~al.(2020)Wang, Li, Khabsa, Fang, and Ma]{wang2020linformer}
Sinong Wang, Belinda~Z Li, Madian Khabsa, Han Fang, and Hao Ma.
\newblock Linformer: Self-attention with linear complexity.
\newblock \emph{arXiv preprint arXiv:2006.04768}, 2020.

\bibitem[Wang and Zhang(2024)]{wang2024osffnet}
Yang Wang and Tao Zhang.
\newblock Osffnet: Omni-stage feature fusion network for lightweight image super-resolution.
\newblock In \emph{Proceedings of the AAAI Conference on Artificial Intelligence}, 2024.

\bibitem[Wang et~al.(2004)Wang, Bovik, Sheikh, and Simoncelli]{wang2004image}
Zhou Wang, Alan~C Bovik, Hamid~R Sheikh, and Eero~P Simoncelli.
\newblock Image quality assessment: from error visibility to structural similarity.
\newblock \emph{IEEE Transactions on Image Processing}, 2004.

\bibitem[Yang et~al.(2022)Yang, Wang, Zhang, Zhang, Wei, Lin, and Yuille]{yang2022lite}
Chenglin Yang, Yilin Wang, Jianming Zhang, He Zhang, Zijun Wei, Zhe Lin, and Alan Yuille.
\newblock Lite vision transformer with enhanced self-attention.
\newblock In \emph{Proceedings of the IEEE/CVF Conference on Computer Vision and Pattern Recognition}, 2022.

\bibitem[Zamfir et~al.(2024)Zamfir, Wu, Mehta, Zhang, and Timofte]{zamfir2024see}
Eduard Zamfir, Zongwei Wu, Nancy Mehta, Yulun Zhang, and Radu Timofte.
\newblock See more details: Efficient image super-resolution by experts mining.
\newblock In \emph{Forty-first International Conference on Machine Learning}, 2024.

\bibitem[Zamir et~al.(2022)Zamir, Arora, Khan, Hayat, Khan, and Yang]{zamir2022restormer}
Syed~Waqas Zamir, Aditya Arora, Salman Khan, Munawar Hayat, Fahad~Shahbaz Khan, and Ming-Hsuan Yang.
\newblock Restormer: Efficient transformer for high-resolution image restoration.
\newblock In \emph{Proceedings of the IEEE/CVF Conference on Computer Vision and Pattern Recognition}, 2022.

\bibitem[Zhang et~al.(2017)Zhang, Zuo, Gu, and Zhang]{zhang2017learning}
Kai Zhang, Wangmeng Zuo, Shuhang Gu, and Lei Zhang.
\newblock Learning deep cnn denoiser prior for image restoration.
\newblock In \emph{Proceedings of the IEEE Conference on Computer Vision and Pattern Recognition}, 2017.

\bibitem[Zhang et~al.(2018{\natexlab{a}})Zhang, Zhou, Lin, and Sun]{zhang2018shufflenet}
Xiangyu Zhang, Xinyu Zhou, Mengxiao Lin, and Jian Sun.
\newblock Shufflenet: An extremely efficient convolutional neural network for mobile devices.
\newblock In \emph{Proceedings of the IEEE Conference on Computer Vision and Pattern Recognition}, 2018{\natexlab{a}}.

\bibitem[Zhang et~al.(2018{\natexlab{b}})Zhang, Li, Li, Wang, Zhong, and Fu]{zhang2018image}
Yulun Zhang, Kunpeng Li, Kai Li, Lichen Wang, Bineng Zhong, and Yun Fu.
\newblock Image super-resolution using very deep residual channel attention networks.
\newblock In \emph{Proceedings of the European Conference on Computer Vision}, 2018{\natexlab{b}}.

\bibitem[Zhang et~al.(2021)Zhang, Chen, Chen, Deng, Xu, and Wang]{zhang2021data}
Yiman Zhang, Hanting Chen, Xinghao Chen, Yiping Deng, Chunjing Xu, and Yunhe Wang.
\newblock Data-free knowledge distillation for image super-resolution.
\newblock In \emph{Proceedings of the IEEE/CVF Conference on Computer Vision and Pattern Recognition}, 2021.

\end{thebibliography}
}


\end{document}